# Exploring the Potential of Citiverses for Regulatory Learning

Isabelle Hupont[1]
European Commission Joint Research Centre
C. Inca Garcilaso, 3, 41092 Sevilla, Spain
https://orcid.org/0000-0002-9811-9397
https://www.linkedin.com/in/isabelle-hupont-torres-95747910/

Marisa Ponti
Department of Applied Information Technology
Faculty of Science and Technology
University of Gothenburg, Sweden
Forskningsgången 6
417 56 Göteborg
Corresponding Author
Mobile 0046 736213051
Marisa.ponti@ait.gu.se
http://orcid.org/0000-0003-4708-4048
https://www.linkedin.com/in/pomar/

Sven Schade
European Commission Joint Research Centre
Via E. Fermi, 2749, 21027 Ispra VA, Italy
https://orcid.org/0000-0001-5677-5209
https://www.linkedin.com/in/sven-schade-7359bb10/

## Abstract

Citiverses hold the potential to support regulatory learning by offering immersive, virtual environments for experimenting with policy scenarios and technologies. This paper proposes a science-for-policy agenda to explore the potential of citiverses as experimentation spaces for regulatory learning, grounded in a consultation with a high-level panel of experts, including policymakers from the European Commission, national government science advisers and leading researchers in digital regulation and virtual worlds. It identifies key research areas, including scalability, real-time feedback, complexity modelling, cross-border collaboration, risk reduction, citizen participation, ethical considerations and the integration of emerging technologies. In addition, the paper analyses a set of experimental topics, spanning transportation, urban planning and the environment/climate crisis, that could be tested in citiverse platforms to advance regulatory learning in these areas. The proposed work is designed to inform future research for policy and emphasizes a responsible approach to developing and using citiverses. It prioritizes careful consideration of the ethical, economic, ecological and social dimensions of different regulations. The paper also explores essential preliminary steps necessary for integrating citiverses into the broader ecosystems of experimentation spaces, including test beds, living labs and regulatory sandboxes.

---

[1] Isabelle Hupont is currently Scientific Advisor at the Spanish Ministry of Science, Innovation and Universities, Paseo de la Castellana, 162, Madrid, Spain. Email: isabelle.hupont@ciencia.gob.es

## 1. INTRODUCTION

Regulatory learning refers to the collection and application of evidence or knowledge relevant to current or future regulatory policies. This knowledge is often generated through the process of experimenting with innovative solutions (Kert, Vebrova, & Schade, 2022). Such learning may stem from overcoming regulatory obstacles in practice or identifying unforeseen risks posed by emerging technologies. The mechanisms for regulatory learning vary and can occur in diverse experimentation settings. Top-down approaches are driven by regulators seeking evidence for new measures, while bottom-up regulatory learning emerges organically from innovation activities within experimentation spaces.

Experimentation spaces, such as test beds, living labs and regulatory sandboxes, are central to regulatory learning. Each offers unique features (Kert, Vebrova, & Schade, 2022): test beds provide technical infrastructure to test technology performance and requirements; living labs emphasize co-creation and early-stage testing of innovative solutions in real-world conditions; and sandboxes allow experimentation under modified regulatory conditions to enhance legal certainty. These spaces are increasingly important as the European Union (EU) rolls out its ecosystem of digital regulations, including the Artificial Intelligence Act (European Parliament, 2024b), the Digital Services Act (European Parliament, 2022b), the Digital Markets Act (European Parliament, 2022a) and the Interoperable Europe Act (European Parliament, 2024a). Policymakers and industry call for robust learning practices to navigate this ecosystem without creating undue barriers.

Within this digital policy landscape, the EU has also unveiled its Web 4.0 and virtual worlds strategy (European Commission, 2023a), envisioning persistent, immersive, social and interactive environments that integrate physical and digital realities (Hupont et al., 2023; Cachia et al., 2024). Among these, *citiverses* have emerged as a key innovation. Ponti et al. (2025) define citiverses as "*interconnected and distributed hybrid and virtual worlds representing, and synchronized with, their physical counterparts. It offers new (administrative, economic, social, policy-making, and cultural) virtual goods/services/capabilities to city and community actors such as citizens, represented as digital avatars*". Citiverses replicate real-world cities, placing community actors (citizens, public servants, policymakers and local authorities) at the centre of interactive, immersive urban simulations (European Commission, 2024b).

Urban environments are dynamic and unpredictable, exposing the limits of traditional policy approaches that rely on control and prediction (Calzati & van Loenen, 2023; Mueller, 2020). However, despite growing interest, actual implementation of citiverses remains limited due to significant technical, social and governance challenges – such as interoperability, digital identity and jurisdictional complexity – poorly addressed by current literature (ITU, 2025). This paper argues that citiverses—interactive, hybrid models of cities—can serve as regulatory learning environments, reframing urban governance as a collaborative and experimental process. By enabling controlled experimentation with technologies, policies and governance models, citiverses can help regulators understand the risks, barriers and real-world impacts of emerging technologies (Madiega & Van de Pol, 2022). We advance this argument by proposing a science-for-policy agenda grounded in European Union (EU) and Organisation for Economic Co-operation and Development (OECD) frameworks, literature on experimental governance, and a Delphi-inspired consultation with high-level experts (policymakers, science advisers and leading researchers). We also outline steps for integrating citiverses into the broader ecosystem of test beds, living labs and regulatory sandboxes, and identify opportunities and challenges for using citiverses to strengthen regulatory learning.

## 2. LITERATURE REVIEW ON REGULATORY EXPERIMENTATION

### 2.1 Regulatory experimentation and urban governance

Academic literature on regulatory experimentation remains relatively limited despite its growing policy relevance. Bennear and Wiener (2019) and Fenwick et al. (2016) call for a shift from static to adaptive regulation, distinguishing unplanned approaches (e.g., crisis responses, and ad-hoc reviews) from planned ones (e.g., periodic reviews, adaptive licensing). Zetzsche et al. (2017) propose a spectrum of strategies for fostering innovation within the financial sector, ranging from permissiveness (through case-by-case approvals or special charters) to structured experimentation and entirely new regulatory frameworks. They introduce the concept of "smart regulation", suggesting a more nuanced and dynamic approach to regulatory design.

There is now broad consensus that rapid technological change requires more agile regulatory approaches (OECD, 2024; European Commission, 2023a). Governments need better evidence to evaluate emerging technologies and reconcile innovation with public safety (Bryan & Teodoridis, 2024). Regulatory experimentation is increasingly seen as a way to generate such evidence, enabling proactive risk and uncertainty identification, and responsible innovation (CRI, 2021). The OECD's *Recommendation for Agile Regulatory Governance* (OECD, 2021) and subsequent reports (OECD, 2022; OECD 2024) highlight experimentation as a key instrument for anticipatory regulation, yet its adoption remains uneven across sectors and jurisdictions. Most experimentation to date has been in financial services, mobility and energy, often in the form of sandboxes. The European Commission's Joint Research Centre found that by early 2023, 12 EU Member States had adopted or were developing regulatory experimentation initiatives, with 3 more considering adoption (Gangale et al., 2023). This signals a shift toward embedding experimentation in the *EU's law toolbox*, but many opportunities remain underexploited (Council of Europe, 2020).

### 2.2 Experimentation spaces and their limits

Experimentation spaces—including test beds, living labs and regulatory sandboxes—provide structured environments for learning about new technologies and policies, each emphasizing different aspects of innovation. Test beds focus on controlled technical trials to verify system performance; living labs bring together stakeholders to co-create and evaluate solutions in real-world conditions; and regulatory sandboxes provide a safe legal framework for innovators to trial solutions under modified rules. These approaches have been instrumental in advancing innovation governance, but they remain constrained by geography, infrastructure and sectoral silos (Kert, Vebrova & Schade, 2022).

Current spaces have key limitations. Their local scope often prevents cross-sector or cross-border learning and their feedback cycles are tied to real-world timelines, which can slow adaptation. Many lack the ability to model complex urban interdependencies or to safely test high-stakes scenarios such as public health crises or climate emergencies. Citizen participation, when present, tends to be small-scale and highly localized, limiting the representativeness of insights gained (Engels et al., 2019).

Citiverses combine the strengths of existing infrastructures while overcoming their limitations: they enable scalable, immersive simulations that integrate diverse datasets (environmental, social, economic), support cross-border and cross-sector experimentation, and can deliver real-time feedback. Their virtual nature eliminates the need for expensive physical setups and allows policymakers to explore multiple "what-if" scenarios without real-world risks or disruptions. Moreover, citiverses can actively involve citizens via avatars, visualizing proposed interventions in ways that promote transparency, trust and inclusive decision-making. This capacity to combine systemic modelling, social participation and safe-fail testing makes them uniquely suited for anticipatory and collaborative urban governance. Supplementary Table 1 in the Appendix compares these spaces in detail with citiverses.

## 2.3 Experimental governance and urban planning

Urban experiments have long been used by cities worldwide to test solutions to address complex sustainability challenges (Bulkeley & Castán Broto, 2013; Bulkeley et al., 2019; Evans & Karvonen, 2014). They often take the form of living labs, testbeds or innovation districts, and share common features such as co-creation, place-based learning and stakeholder collaboration (Evans & Karvonen, 2014). Local governments like city councils frequently lead these initiatives, leveraging their resources and authority to foster partnerships with private actors and academia (van der Heijden, 2018; Eneqvist & Karvonen, 2021). This tradition of experimentation makes cities natural early adopters of citiverse-based approaches. By embedding immersive simulations into urban governance processes, citiverses could extend these initiatives beyond physical pilots, enabling large-scale modelling of urban interventions before implementation. Having established the conceptual and empirical limitations of existing experimentation spaces, we now turn to the European policy context to examine why citiverses are gaining momentum as part of the EU's digital agenda.

## 3. EUROPEAN POLICY BACKROUND ON REGULATORY LEARNING

### 3.1 Beyond traditional methods

EU policymakers are increasingly advocating for more flexible, operational, human-centric, collaborative and adaptive approaches to regulation, aiming to balance innovation with societal well-being and fundamental rights. The Council of the European Union (2020) has formally committed to making EU legislation more forward-looking and innovation-friendly by incorporating tools like regulatory sandboxes. These initiatives reflect a growing recognition that traditional regulatory learning methods –quasi-experiments and expert-driven assessments- are insufficient for today's fast-moving technological and social landscapes.

Traditional approaches risk oversimplifying complex socio-technical systems, failing to capture emergent behaviours, feedback loops and cross-sector interdependencies (Mueller, 2020; Kert, Vebrova & Schade, 2022). Experimentation spaces such as test beds, living labs and regulatory sandboxes, represent significant advancements over these traditional methods but remain constrained by local scope, slow feedback cycles and limited citizen engagement (Engels et al., 2019). These limitations point to the need for next-generation infrastructures capable of real-time simulation, systemic modelling and inclusive engagement at scale.

## 3.2 Citiverses in the EU policy context

Among the most promising candidates to meet this need are citiverses: immersive, large-scale virtual environments that integrate simulation-driven engagement, digital twin infrastructure, AI-based modelling and real-time feedback. Building on the strengths of test beds, living labs and sandboxes, citiverses introduce an entirely new level of systemic, immersive experimentation. They enable cost-effective, scalable, interoperable simulations of complex, cross-sector and high-stakes policy scenarios. A key advantage lies in their capacity for safe-fail testing in controlled virtual environments, allowing policymakers to explore regulatory options, anticipate unintended effects, and refine measures before real-world implementation.

These environments also foster multi-stakeholder interaction through avatars, enabling regulators, innovators and citizens to co-exist in the same simulated space. Regulators can assess societal impacts in real time, businesses can trial novel solutions within simulated legal frameworks, and citizens can experience and respond to interventions –building trust, inclusion and social foresight. By supporting cross-sector and cross-border collaboration, citiverses offer a shared platform for aligning regulatory practices and addressing transnational challenges.

The EU has already recognized citiverses' potential, designating them as a "*flagship project of public interest*". Its vision highlights their role in optimizing spatial planning and management while respecting social, sustainability and cultural heritage dimensions (European Commission, 2023b; Global Cities Hub, 2023). Recent initiatives, including the CitCOM[2] Testing and Experimentation Facility (TEF) and the citiverse EDIC (European Digital Infrastructure Consortium), aim to provide shared cloud-based, AI-empowered platforms and digital twins for smart cities across Europe. These steps are aligned with the EU's Digital Decade programme[3] and its ambition to employ simulations for next generation urban solutions. In its Communication on Boosting Startups and Innovation in Trustworthy Artificial Intelligence (European Commission, 2024a), the European Commission recommends the use of AI in smart cities applications for citiverses by employing simulations of scenarios "*such as the impact of changing traffic conditions on air quality, decarbonisation and congestion and more broadly on greening cities*".

Despite these advances, citiverses are at an early stage of implementation. Effective and responsible use for regulatory learning will require a clear science-for-policy agenda to guide development, address technical and governance challenges, and ensure alignment with European values and public objectives.

## 4. RESEARCH APPROACH

### 4.1 Survey methodology and participants

In May 2025, we conducted a single-round, Delphi-inspired online survey to identify key research areas, questions and applications for citiverses as spaces for regulatory learning. 24 experts were invited from high-level institutions, including European Commission Directorates-General and national government science advisers, and 10 participated. The study prioritized expert judgment quality over statistical generalizability, a common approach in policy Delphi studies for emerging fields where empirical knowledge is limited (Hsu & Sandford, 2007; Pahwa, Cavanagh & Vanstone, 2023).

---

[2] https://citcom.ai/

[3] The Digital Decade programme was established with EU Decision, 2022/2481.

The final sample offered a diverse mix of institutional perspectives, including three science advisers to the Spanish government, one to the UK government, three researchers from the European Commission's Joint Research Centre, and three policymakers from the European Commission working on digital technologies. This diversity provided a rich blend of national and European-level viewpoints. To account for the participants' limited availabilities, an asynchronous online survey was chosen over focus groups to maximise feasibility. The survey was deployed using EUSurvey and is publicly available at https://ec.europa.eu/eusurvey/runner/SurveyCitiverse.

Supplementary Figure 1 (Appendix) summarises participants' self-assessed expertise in regulatory learning (x-axis), virtual worlds (y-axis) and citiverses (circle size), revealing complementary strengths. Overall citiverse expertise is low (1–3 on a 5-point Likert scale), consistent with the novelty of the concept.

## 4.2 Questionnaire design

The questionnaire was custom-built to gather expert perceptions on this novel topic, drawing on the preceding literature and policy reviews (Sections 2 and 3) to ensure relevance. A non-standardized instrument was developed because none existed for the novel citiverse concept, and the study's aim was exploratory data collection rather than measuring an established stable construct. As the notion of citiverse is nascent, a rigid questionnaire would have been premature and potentially limiting. Instead, a flexible design allowed us to capture nuanced or unexpected perspectives, which can inform the development of more structured tools in the future.

As the survey did not collect sensitive personal data under the European General Data Protection Regulation (GDPR), ethical approval was not required. Informed consent was obtained from all participants.

## 4.3 Limitations

This study has several limitations that should be considered when interpreting the results. The availability of senior level experts (10 of 24 invitees) posed feasibility challenges. Some invitees declined due to limited familiarity with the citiverse concept, highlighting both the novelty of the topic and the difficulty of recruiting knowledgeable experts.

While the panel size falls within the accepted range for exploratory Delphi studies (Okoli & Pawlowski, 2004; Hsu & Sandford, 2007) and participants were carefully selected for their expertise, a larger and more diverse panel could have provided additional perspectives and strengthened the robustness of the findings. Results should therefore be seen as indicative rather than statistically generalisable.

The analyses in Section 5 (e.g. violin plots, boxplots) are intended to illustrate patterns of convergence and divergence in expert views, not to support inferential claims. Finally, because the consultation was conducted in a European policy context, the findings cannot be directly generalised to non-European regulatory regimes.

# 5. RESULTS

This section presents the findings from the expert consultation, covering research priorities, experimental topics, perceived barriers and design recommendations. The anonymised dataset is available upon reasonable request to the corresponding author.

## 5.1 Analysis of research priorities for regulatory learning in citiverses

To identify which Research Areas (RAs) experts consider most relevant for advancing regulatory learning in citiverses, participants were asked in question Q7 of the survey to rate 26 predefined items on a 5-point Likert scale (0=*not important at all*; 4=*absolutely essential*). These items spanned technological, social, ethical and regulatory dimensions, and were derived from the literature and policy reviews in Sections 2 and 3, to ensure alignment with current academic debates and institutional agendas.

Participants could also suggest additional RAs through an open-ended item (Q8). Only 3 respondents did so, and two of their remarks related not to new areas but to feasibility and prioritisation — for example, noting that lower-rated RAs should not be seen as irrelevant but as less urgent, and that initial experimentation should start in low-risk sectors. These comments anticipate later analysis, as they highlight the importance of combining perceived importance with levels of agreement when identifying robust priorities.

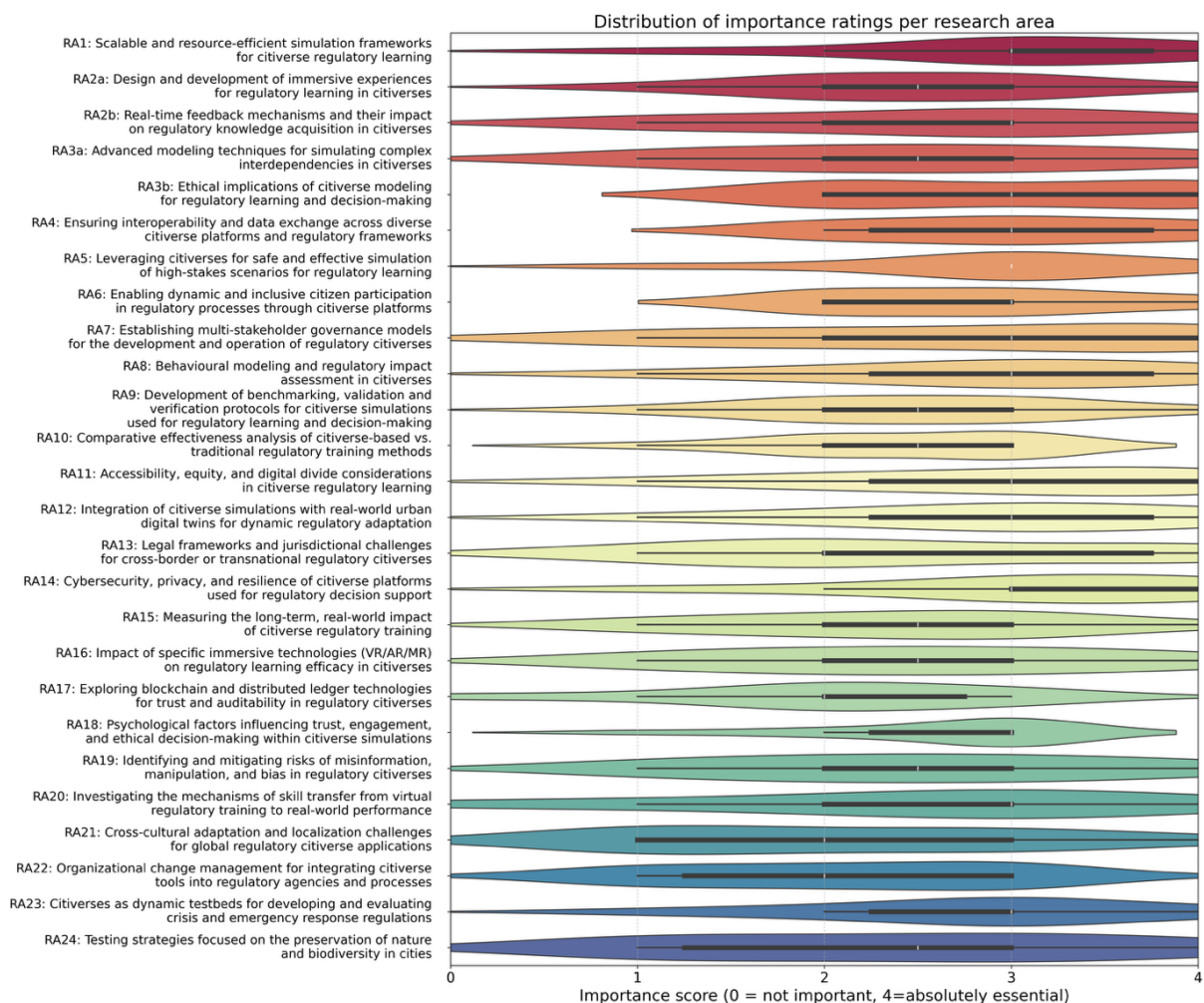

*Figure 1. Distribution of importance ratings per Research Area (RA).*

Figure 1 displays the distribution of ratings per RA as violin plots. 14 RAs (54%) received a median score between 3 and 4, and 20 (77%) had a mean score ≥2.5, indicating broad recognition of their relevance. This suggests that citiverse-based approaches are widely seen as important for regulatory innovation.

Several RAs stand out for their ratings patterns. RA18 received a median of 3, with responses tightly clustered, indicating strong agreement on its importance. In contrast, RA20 shows a wider dispersion of scores (from 0 to 4), reflecting more divergent views. RA1 displays a bimodal distribution—clusters at both low and high ends—suggesting polarized perceptions of its relevance.

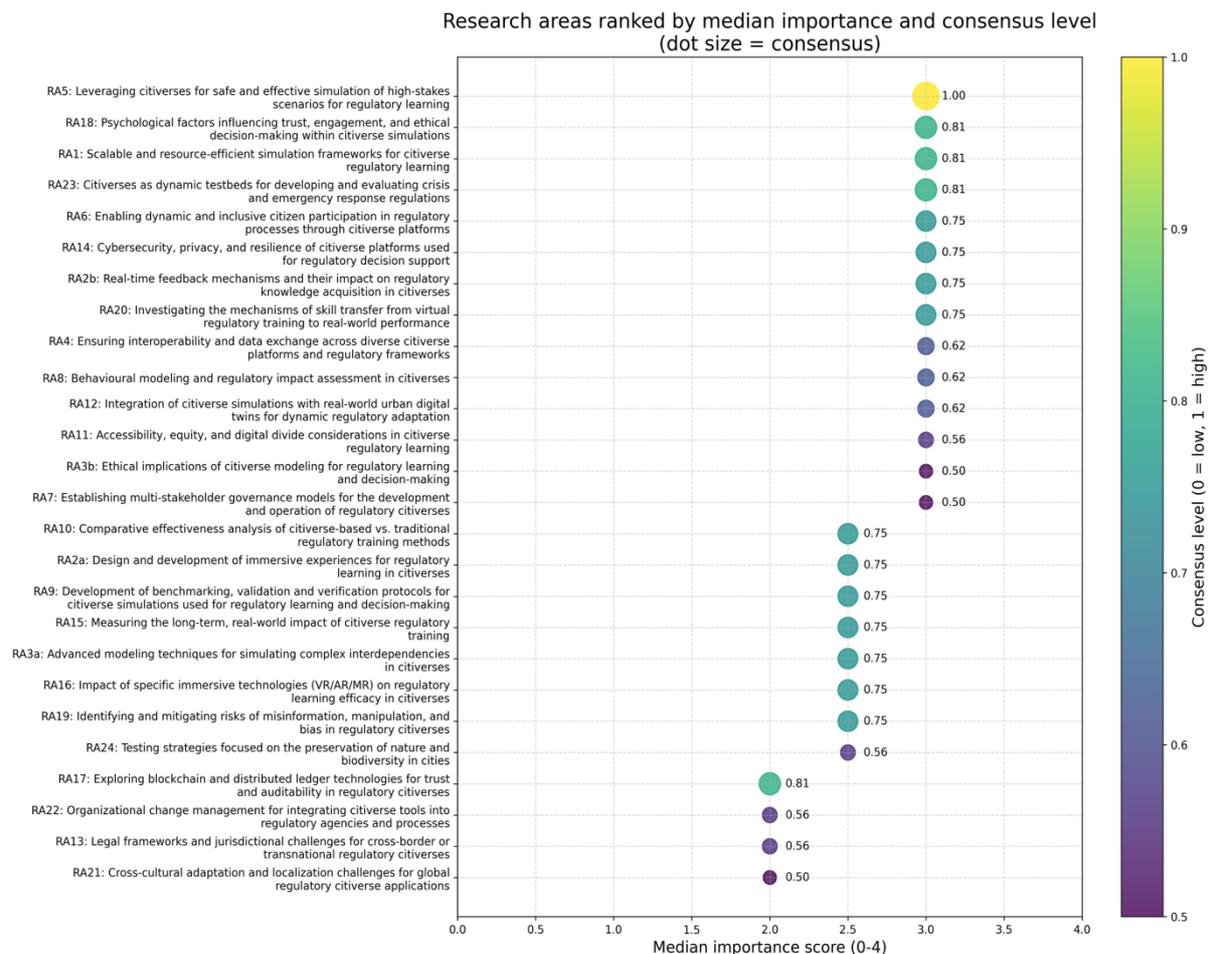

*Figure 2. Research areas ranked by median importance and consensus level.*

To capture both perceived importance and the degree of agreement, we ranked RAs using two measures: the median score, reflecting central importance, and a consensus index (inverse normalized Shannon entropy; Tastle & Wierman, 2007), ranging from 0 (no agreement) to 1 (full agreement). This combination captures not only how important an RA was judged but also how strongly experts converged around that judgment. Figure 2 plots each RA as a circle, positioned by its median (x-axis) and sized/coloured by agreement level.

Top-ranked RAs combined high importance (median 3.0) with strong consensus (≥ 0.81):

- RA5: *Leveraging citiverses for safe and effective simulation of high-stakes scenarios for regulatory learning* (median 3.0, consensus 1.00).

- RA18: *Psychological factors influencing trust, engagement and ethical decision-making within citiverse simulations* (median 3.0, consensus 0.81).

- RA1: *Scalable and resource-efficient simulation frameworks for citiverse regulatory learning* (median 3.0, consensus 0.81).

- RA23: *Citiverses as dynamic testbeds for developing and evaluating crisis and emergency response regulations* (median 3.0, consensus 0.81).

These RAs highlight a shared perception of the strategic value of immersive simulation environments for planning, training and regulatory decision-making. RA5, RA1 and RA23 converge on simulation frameworks, reflecting a strong belief in the centrality of immersive environments for high-stakes scenario testing, risk mitigation and policy prototyping. RA18 complements this cluster with a human-centered focus, emphasising the psychological dynamics of trust, engagement and ethical decision-making. Its top-tier ranking underscores that technical systems must be evaluated not only for functionality, but also for their credibility and ethically grounded interactions with users. Together, these four RAs form a coherent set of research priorities, grounded in strategic foresight, public trust and real-world applicability. Their combination of high importance and high consensus suggests they represent particularly robust entry points for early policy action and research investment.

At the other end, RAs with both low median and low consensus include RA21 (*cross-cultural adaptation and localization challenges for global regulatory citiverse applications;* median 2.0, consensus 0.50), RA13 (*legal frameworks and jurisdictional challenges for cross-border or transnational regulatory citiverses;* median 2.0, consensus 0.56) and RA22 (*organisational change management for integrating citiverse tools into regulatory agencies and processes;* median: 2.0, consensus 0.56). These may be considered longer-term challenges or less feasible for early implementation, which aligns with Q8 comments that lower-rated topics reflected prioritisation rather than irrelevance.

One notable case is RA17 (*blockchain and distributed ledger technologies*). Though rated lower (median 2.0), it shows high consensus (0.81), consistent with recent literature reassessing blockchain's near-term policy utility (Joshi et al., 2023) and pointing to emerging alternatives (e.g., W3C Verifiable Credentials[4] and the EU Digital Identity Wallet[5]).

The remaining Q8 contribution specifically mentioned "*maturity, sustainability and avoidance of technological lock-in in citiverse platforms*". This overlaps with barriers further discussed in Section 5.3 and reinforces that the predefined list of RAs was broadly comprehensive, with open-ended suggestions serving mainly to stress feasibility, prioritisation and implementation issues.

## 5.2. Use cases: Prioritising experimental topics for regulatory learning in citiverses

---

[4] https://www.w3.org/TR/vc-data-model-2.0/
[5] https://ec.europa.eu/digital-building-blocks/sites/display/EUDIGITALIDENTITYWALLET/EU+Digital+Identity+Wallet+Home

To complement the analysis of research priorities in Section 5.1 and prepare the ground for the subsequent discussion on barriers and design recommendations (Section 5.3), the survey asked in question Q9 for participants to prioritise 12 potential experimental topics for citiverse-based regulatory experimentation. These topics were selected not as an exhaustive list, but as illustrative examples of current pressing urban challenges visible in policy agendas - such as climate adaptation, energy transition, mobility and social equity (Alaassar et al., 2020; European Parliament, 2024c).

The goal was to encourage reflection and contextualise the broader research agenda in more applied and actionable domains. Concrete examples also helped participants be better equipped to assess the feasibility and limitations of citiverse experimentation, which were further explored in questions Q10 and Q11 (Section 5.3).

Responses were analysed using the Borda count method (Emerson, 2013), a well-established approach in preference aggregation. Rankings were converted into Borda scores from 1 to 12 and then averaged across participants. Figure 3 displays the distribution of Borda scores per experimental topic in the form of boxplots.

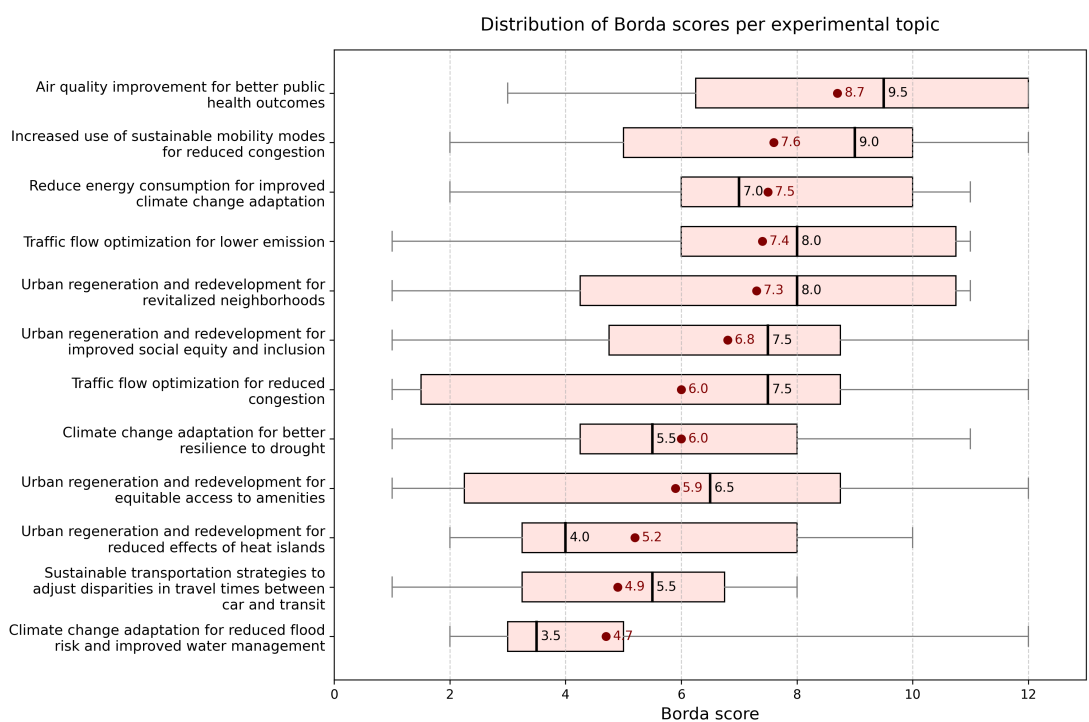

*Figure 3. Distribution of Borda scores per experimental topic. Boxplots show variability in experts' rankings; dark red circles indicate mean score.*

The five most prioritised experimental topics were: *Air quality improvement for better public health outcomes* (mean = 8.7; median = 9.5), *Increased use of sustainable mobility modes for reduced congestion* (mean = 7.6; median = 9.0), *Reduce energy consumption for improved climate change adaptation* (mean = 7.5; median = 7.0), *Traffic flow optimization for lower emission* (mean = 7.4; median = 8.0) and *Urban regeneration and redevelopment for revitalized neighborhoods* (mean = 7.3; median = 8.0). While all twelve Q9 topics intentionally fall within environmental quality, climate action and urban transformation, these results indicate relative prioritisation within this predefined set. The prominence of these five likely reflects both their salience in current policy agendas and their perceived tractability for early citiverse experimentation.

Rather than offering definitive conclusions, this exercise illustrates the real-world challenges experts view as promising entry points for regulatory learning in citiverses within the scope of Q9's topic list. It also reflects current policy sensitivities, offering useful guidance for strategically targeting citiverse deployments in support of pressing societal goals.

## 5.3 Perceived barriers and design recommendations for citiverse deployment in regulatory learning

Two open-ended questions (Q10 and Q11) invited participants to identify key barriers to using citiverses for regulatory learning and to propose design recommendations, respectively. We have grouped responses into six thematic categories aligned with the Research Areas (RAs). Illustrative quotes are integrated below, with a synthesis provided in Table 1.

**Data governance and integration**. Respondents pointed to persistent gaps in data ownership, access and protection, citing "*data protection and proper access mechanisms*" and "*lack of transparency and auditability*" as relevant barriers. They also stressed the challenge of integrating heterogenous datasets: platforms should "*integrate data from multiple sources and channels in a way that fits the specific needs of citiverse contexts*". These comments suggest the importance of addressing both legal and technical aspects of data management. These concerns relate to several research areas previously identified in the study, including RA13 (legal frameworks and jurisdictional challenges), RA14 (cybersecurity, privacy and resilience) and RA17 (trust and auditability through distributed ledger technologies). Design recommendations include implementing explicit consent and access protocols, incorporating auditability mechanisms, and ensuring that data integration strategies are adapted to the specific requirements of the regulatory domain.

**Technological maturity and implementation feasibility**. Several participants noted the limited maturity of current platforms, and difficulties in integrating real-time sensor data and complex modelling. The problem of fragmented and non-interoperable platforms was also highlighted. These observations align with RA1 (scalable frameworks), RA3a (advanced modelling techniques), RA4 (interoperability) and RA12 (integration with digital twins). Suggested solutions include developing lightweight, modular platforms, prioritising technical robustness, and start piloting in low-risk domains to evolve models iteratively.

**Institutional readiness and skills gaps.** Limited digital competences and resistance to change within public administration were emphasised: "*There is a lack of professional profiles and digital skills in the public service to manage such tools*". A participant pointed to "*the procedures established in the administration*" as a source of inertia. These reflections resonate with RA7 (governance models) and RA22 (organisational change management). Recommendations include targeted training of public servants, fostering interdisciplinary collaboration and co-design with regulators to align citiverse tools with existing workflows.

**Political and social acceptance**. Participants expressed skepticism about political support and public trust. "*The main barrier would be political will*", noted one respondent. Others warned against "*vague promises not supported by a clear value proposition*", which may result in user disappointment. These views are relevant to RA6 (inclusive citizen participation), RA7 (multi-stakeholder governance) and RA18 (psychological factors influencing trust). Suggested measures include transparent communication of benefits and limitations, early stakeholder engagement, and piloting in low-risk domains to build legitimacy.

**Sustainability and long-term risks**. Concerns were raised about the risk of technological lock-in, high initial setup costs and uncertain return on investment. These issues were raised in the context of deploying citiverse platforms within fast-evolving digital ecosystems, where maintaining flexibility and cost-efficiency is critical. Although not directly mapped to any of the predefined research areas, these observations point to the importance of addressing sustainability from both an economic and technological standpoint. Recommendations include promoting open architectures—which may reduce dependence on proprietary systems—and conducting strategic scoping, lifecycle assessments and risk analyses before large-scale implementation and investment.

**Evidence and validation**. Participants raised questions about the ecological validity of citiverse-generated insights and the challenge of transferring simulation results to real-world policy. One respondent remarked: "*Evaluating the accuracy and showing use cases will be key for policymakers to see the value of and engage with citiverses*". These concerns are directly related to RA5 (simulation of high-stakes scenarios), RA8 (regulatory impact assessment), RA9 (validation and benchmarking protocols), RA15 (real-world impact of regulatory training) and RA20 (skill transfer from simulation to real-world settings). Recommendations include measurable indicators, explainable interfaces and rigorous evaluation protocols to ensure transferability of results to real-world contexts

The barriers and recommendations outlined above suggest that participants see the deployment of citiverses as contingent on a complex interplay of legal, technical, institutional and social factors. While further interpretation of these findings is developed in Section 6, this analysis provides an initial mapping of the perceived prerequisites for the successful use of citiverses in regulatory learning contexts.

| Category | Perceived barriers | Design and implementation recommendations | Related RA(s) |
|---|---|---|---|
| **Data governance and integration** | - Unclear data ownership<br>- Need for data protection, data quality and data access frameworks<br>- Integrating heterogeneous data sources (e.g. real-time, administrative, citizen-generated) in ways tailored to specific citiverse policy contexts<br>- Lack of transparency and auditability mechanisms | - Embed clear consent and data access protocols<br>- Develop data architectures that securely integrate diverse sources and reflect the specific requirements of citiverse regulatory domains<br>- Design platforms with built-in auditability mechanisms | RA13, RA14, RA17 |

| Category | Perceived barriers | Design and implementation recommendations | Related RA(s) |
|---|---|---|---|
| **Technological maturity and implementation feasibility** | - Current platforms are often not mature enough, struggling to go beyond simple visualizations and integrate complex modelling or real-time data<br>- Fragmentation of platforms<br>- Lack of interoperability with existing tools and systems | - Develop modular, scalable and lightweight platforms and modelling frameworks<br>- Prioritise technical robustness over hype<br>- Start with simple, low-risk domains and iteratively evolve models | RA1, RA3a, RA4, RA12 |
| **Institutional readiness and skills gaps** | - Limited competences, skills and professional profiles in public administration<br>- Resistance due to procedural lock-in or digital inertia<br>- Misalignment between citiverse functionality and organisational workflows | - Co-design citiverses with regulators and civil servants<br>- Train staff in relevant digital and participatory skills<br>- Foster interdisciplinary teams to bridge governance, design and technical needs | RA7, RA22 |
| **Political and social acceptance** | - A lack of political will could block the use of the technology even if available<br>- Risk of public disinterest or resistance<br>- Unrealistic promises leading to disappointment | - Provide clear value propositions<br>- Define expectations by communicating transparently what citiverses can and cannot do<br>- Involve stakeholders in early stage design<br>- Use low-risk pilots to build trust | RA6, RA7, RA18 |
| **Sustainability and long-term risks** | - Risk of technological lock-in<br>- Uncertain cost-effectiveness and return on investment<br>- Risk of sunk investments in fast-changing innovation landscape | - Use open architectures<br>- Perform lifecycle and risk assessments<br>- Strategic scoping of citiverse utility before development | *Potential new RA* |
| **Evidence and validation** | - Uncertainty about the ecological validity of insights generated from citiverse simulations<br>- Difficulty in transferring simulation insights to real-world policy<br>- Heterogeneity in regulatory environments | - Ensure clearly scoped experiments with measurable indicators and rigorous evaluation of impact<br>- Design explainable and interpretable interfaces<br>- Enable local calibration and contextualisation of models<br>- Ground claims in scientific evidence | RA5, RA8, RA9, RA15, RA20 |

*Table 1. Summary of barriers and design recommendations identified by participants for citiverses as regulatory learning spaces.*

## 6. DISCUSSION: PROPOSED SCIENCE-FOR-POLICY RESEARCH AGENDA

Building on the expert consultation, this section develops a science-for-policy agenda for citiverses as regulatory learning spaces. To ensure relevance and feasibility, the agenda concentrates on the four research areas that were top-ranked in Section 5.1 (RA1, RA5, RA18 and RA23). These areas combined high perceived importance with strong consensus among experts, making them robust entry points for early policy action and research investment.

Our proposed agenda is structured around two core pillars: (i) building the foundational simulation engine (linked to RA1 and RA23), and (ii) ensuring human-centric validity and trust (linked to RA5 and RA18). Each pillar brings together research areas that address the dual challenges of technological readiness and socio-political acceptance that emerged from our analysis. For each RA included in each pillar we propose associated research questions, a brief rationale, and the specific barrier(s) they address.

## 6.1 Pillar 1: Building the foundational simulation engine

The expert consultation underlined that the primary value of citiverses lies in their simulation capabilities, suggesting a conflation between the nascent idea of a "citiverse" and the more established concept of a "digital twin". Before citiverses can support complex regulatory learning, their underlying technical engine must be powerful, efficient and capable of modeling real-world crises. This pillar combines two RAs focused on creating such core technical capacity.

### *Scalable and resource-efficient simulation frameworks for citiverse regulatory learning* (RA1)

Citiverses can enable large-scale simulations of urban environments that would otherwise be too costly or difficult to recreate. This requires moving beyond simple visualizations to test multiple regulatory scenarios simultaneously without disrupting real systems. Examples of research questions and barriers addressed are listed in Table 2.

| Research Question | Rationale | Barrier(s) Addressed |
|---|---|---|
| • How can scalable simulations improve the efficiency of testing regulatory measures? | • Move beyond small-scale tests and enable comprehensive, simultaneous scenario analysis. | • Technological Maturity |
| • What frameworks are needed to ensure these simulations are realistic and actionable? | • Ensure simulation outputs are reliable enough to inform real-world policy. | • Technological Maturity; Evidence & Validation |

*Table 2. Research questions, rationales and barriers addressed for research area: Scalable and resource-efficient simulation frameworks for citiverse regulatory learning.*

### *Citiverses as dynamic testbeds for developing and evaluating crisis and emergency situations* (RA23)

As urban crises grow in complexity, physical field exercises become increasingly costly and risky. By integrating real-time data, citiverses can function as dynamic, risk-free testbeds to simulate cascading failures, model population movements and test response strategies. Table 3 outlines research questions and barriers addressed.

| Research Question | Rationale | Barrier(s) Addressed |
|---|---|---|
| • How can citiverses model the complex, cascading failures of critical infrastructure to a degree of realism not feasible in field exercises? | • Simulate events that are too complex and risky to test with physical exercises. | • Evidence & Validation |

| Research Question | Rationale | Barrier(s) Addressed |
|---|---|---|
| • What is the comprehensive cost-benefit analysis and ROI for a city developing a citiverse testbed? | • Justify the significant investment required and ensure long-term viability. | • Sustainability & Long-Term Risks |
| • What new psychological or ethical risks do citiverses introduce, and what governance approaches are needed to mitigate them? | • Proactively identify and mitigate unintended negative consequences of using immersive simulations. | Data Governance; Political & Social Acceptance |

*Table 3. Research questions, rationales and barriers addressed for research area: Citiverses as dynamic testbeds for developing and evaluating crisis and emergency situations.*

## 6.2 Pillar 2: Ensuring human-centric validity and trust

Even the most sophisticated simulation engine is of little value if its results are not trusted or perceived as legitimate. This pillar addresses the crucial human dimension of citiverses, focusing on the factors that determine their social acceptance and real-world effectiveness.

*Leveraging citiverses for safe and effective simulation of high-stakes scenarios for regulatory learning* (RA5)

By operating in a virtual space, citiverses provide a safe environment to test policies involving significant risks, such as public health crises or infrastructure failures. Their effectiveness, however, depends on translating virtual insights into actionable, real-world policies that are accepted by both decision-makers and the public. Examples are presented in Table 4.

| Research Question | Rationale | Barrier(s) Addressed |
|---|---|---|
| • How can virtual environments effectively simulate these scenarios? | • Ensure high-stakes simulations are sufficiently accurate and comprehensive. | • Evidence & Validation |
| • Which strategies can be applied to translate insights into actionable real-world policies? | • Create a clear pathway from virtual experimentation to tangible policy change. | • Evidence & Validation |

*Table 4. Research questions, rationales and barriers addressed for research area: Leveraging citiverses for safe and effective simulation of high-stakes scenarios for regulatory learning.*

*Psychological factors influencing trust, engagement, and ethical decision-making within citiverse simulations* (RA18)

Citiverses may represent a novel paradigm for participatory governance. However, their effectiveness relies on resolving key psychological and ethical questions to design trustworthy and engaging tools for collaborative policymaking. Citiverses allow the public to participate as digital avatars, which is especially valuable for gauging social acceptance and anticipate unforeseen social impacts. Table 5 illustrates related research questions and barriers.

| Research Question | Rationale | Barrier(s) Addressed |
|---|---|---|
| • How does perceived realism and data transparency influence citizen trust in a simulation's processes and recommendations? | • To understand the core drivers of public trust, which is essential for the legitimacy of any participatory platform. | • Political & Social Acceptance |
| • What is the relationship between a participant's sense of agency (via their digital avatar) and their engagement in policy-making tasks? | • To optimize the user experience to encourage deep and meaningful participation. | • Political & Social Acceptance |
| • How do cognitive biases and digital representation affect decision-making and the perceived fairness of policy impacts? | • To ensure simulations do not inadvertently reinforce existing biases and lead to inequitable policy outcomes. | • Political & Social Acceptance; Evidence & Validation |

*Table 5. Research questions, rationales and barriers addressed for research area: Psychological factors influencing trust, engagement, and ethical decision-making within citiverse simulations.*

## 6.3 Methodological approach to operationalise the agenda

To operationalize the proposed science-for-policy agenda, a cross-cutting methodological approach is required. This approach must capture both the nuanced, qualitative dynamics of human interaction within virtual worlds and the broader, quantitative impacts on policy and governance. We therefore propose two complementary methodological components: (i) experimental prototyping and in-depth case analysis; and (ii) indicators and large-scale impact assessment.

### *Experimental prototyping and in-depth case analysis*

As the implementation of citiverses is still quite limited, research cannot rely solely on observing existing examples. Instead, an active, experimental approach is needed. This involves building and studying prototypes to understand the internal workings of regulatory learning in these immersive environments. This experimental stream can be structured around two key activities:

- **Deep dives into prototypes**: Rather than simply mapping use cases, research should focus on undertaking deep dives into selected experimental prototypes. This requires qualitative methods like in-depth interviews with participating policymakers, urban planners, and citizens; observational analysis of user behavior within simulations; and document analysis of the co-design and development process. This approach provides crucial evidence for understanding how citiverses can be integrated into organizational structures and decision-making workflows.

- **Identifying best practices**: A key goal is to identify and present best-practice examples. Based on a clear set of criteria (e.g., effectiveness of engagement, validity of insights, trustworthiness), these cases can have a powerful communicative effect. They can highlight specific elements of the experimental process—from stakeholder co-design to translating virtual insights into policy—allowing regulators and researchers to identify successful pathways and potential pitfalls.

*Indicators and large-scale impact assessment*

While citiverses are envisioned to support regulatory learning, tools to measure their impact remain scarce. Future research should prioritise the development of dedicated indicators and assessment methodologies, organised around three components:

- **Developing novel indicator sets:** these new indicators should move beyond technical performance to measure outcomes central to the research agenda, such as:

    o The effectiveness of regulatory learning (e.g., time saved in policy development, reduction of real-world implementation risks).
    o The quality and inclusiveness of citizen engagement and the impact on public trust.
    o The cost-benefit ratio of virtual experimentation compared to traditional pilots or high-risk physical exercises.

- **Causal inference with Randomized Controlled Trials (RCTs)**: To move beyond correlation and rigorously establish the causal impact of citiverses, researchers can design RCTs. For instance, participants (citizens or officials) could be randomly assigned to different groups: an intervention group using an immersive citiverse to evaluate a proposed urban development, and a control group reviewing the same proposal using traditional methods like 2D blueprints and reports. Comparing outcomes—such as policy comprehension, social acceptance, or perceived fairness of the proposal— would provide robust causal evidence on whether citiverses are more effective than existing methods for regulatory learning and public engagement.

- **Large-Scale assessment and benchmarking**: Existing benchmarks for smart cities and digital government provide a starting point, but these must be complemented with new instruments. For example, indicators could measure the ecological validity of simulation outcomes by systematically comparing them to real-world data, or assess the transferability of skills gained by public officials in virtual crisis-response exercises to their real-world performance. Such indicators are essential for generating data-driven policy and justifying investment in these advanced digital infrastructures.

## 6.4 Bridging the literature review and the science-for-policy agenda

The proposed science-for-policy agenda responds to the challenges and gaps identified in the literature on regulatory experimentation and experimental governance. While our literature review establishes a clear need for more dynamic and evidence-based regulatory approaches, it also highlights that the effective use of experimentation by regulators is still in its early stages and varies considerably (CRI, 2021; OECD, 2024). Our agenda aims to advance this field by operationalizing citiverses as novel experimentation spaces.

A central theme from the literature is the call for a shift from static to planned adaptive regulation to cope with technological and societal uncertainty (Bennear & Wiener, 2019). Our proposed science-for policy agenda directly addresses this by prioritizing the use of citiverses for simulating high-stakes scenarios (RA5) and evaluating crisis responses (RA23). These research areas provide a tangible mechanism for the "planned adaptive approaches" that scholars advocate for. Instead of relying on ad-hoc reviews or waiting for crises to unfold, citiverses offer a "safe-to-fail" environment where regulators can proactively test and refine policies, building institutional resilience. Furthermore, the literature on urban experimentation emphasizes its role in addressing complex sustainability challenges through place-based, collaborative learning (Bulkeley & Castán Broto, 2013; Evans & Karvonen, 2014). In line with this, our research agenda proposes to investigate the critical human-centric elements of these experiments. By prioritizing research into the psychological factors that influence trust, engagement, and ethical decision-making (RA18), the agenda seeks to understand *how* to make these virtual urban experiments genuinely participatory and legitimate. This directly aligns with the literature's focus on co-creation and stakeholder involvement, offering a pathway to operationalize these concepts in a scalable and immersive digital context.

Finally, a persistent challenge noted in the literature is that governments and regulators often lack the knowledge needed to effectively evaluate new technologies and reconcile innovation with public safety (OECD, 2024). The emphasis on developing scalable and resource-efficient simulation frameworks is a direct response to this knowledge gap. By enabling large-scale, cost-effective simulations, citiverses can serve as powerful engines for regulatory learning, generating the evidence needed to inform policy. This moves beyond the more limited scope of traditional regulatory sandboxes, which are just one of many potential tools (OECD, 2024). The proposed agenda, therefore, seeks to develop a more sophisticated and powerful platform for generating the forward-looking evidence required for agile and innovation-friendly governance in the 21st century.

## 6.5 Aligning the research agenda with the European policy context

Beyond its dialogue with academic literature, the proposed research agenda is fundamentally an instrument for science for policy, designed to directly support and inform the European Union's strategic ambitions in the digital domain. The EU's vision, as articulated in its communications on Web 4.0 and virtual worlds, is not merely to adopt new technologies but to shape them in line with European values and policy priorities (European Commission, 2023b). Our research agenda provides a structured pathway to ensure this vision is realized in a responsible, evidence-based manner.

The European Commission's designation of the citiverse as a "*flagship project of public interest*" and its investment in infrastructure like the citiverse EDIC and the CitCOM Testing and Experimentation Facility (TEF) signal a clear commitment to leveraging these virtual environments for public good. The research agenda directly underpins these initiatives. The focus on scalable and resource-efficient simulation frameworks (RA1) is essential for building the shared, cloud-based platforms the EDIC is tasked with developing. Furthermore, the emphasis on using citiverses as dynamic testbeds for crisis scenarios (RA23) and for simulating complex policy trade-offs—such as the impact of traffic on air quality as recommended by the European Commission (European Commission, 2024a)—provides the methodological foundation needed to make these facilities effective tools for policymaking. In short, the agenda offers the "how" to the EU's "what," translating high-level policy goals into actionable research.

Moreover, the EU's approach to digitalization is consistently framed by a commitment to human-centricity, trust and ethical responsibility. This is where research on the psychological factors influencing trust, engagement, and ethical decision-making (RA18) becomes critically important. While the EU provides the high-level regulatory frameworks, RA18 seeks to understand the on-the-ground conditions necessary for these systems to be perceived as legitimate and fair by citizens. By investigating how to foster trust and ensure ethical interaction within these virtual worlds, the agenda helps ensure that the development of citiverses aligns with the core principles of the "European way" for the digital transition—one that is open, secure and serves people first.

## 7. CONCLUSIONS

We have proposed a science-for-policy agenda to unlock the potential of citiverses as spaces for regulatory learning. Grounded in a high-level expert consultation, the agenda is structured around two core, complementary pillars: building the foundational simulation engine needed to model complex policy scenarios, and ensuring the human-centric ecological validity and trust required for the learning to be socially accepted, legitimate and actionable. By applying this framework to critical domains like transportation and urban planning, we outline a path for generating the forward-looking evidence that modern regulatory learning requires.

While our science-for-policy agenda provides a structured starting point, it is not exhaustive. Further research is needed to build upon the agenda by prioritizing and expanding on the identified challenges and questions. This agenda should be seen as a snapshot in time, as new challenges and gaps are expected to emerge as technologies evolve and institutional practices mature. Continuous work is necessary to keep the agenda current and maintain a productive relationship between research and policy-making. Future work should also begin moving from theoretical discussions to practical experimentation, setting up increasingly complex experiments that specifically probe the unique social, ethical and participatory affordances of citiverses – both within and beyond urban planning scenarios.

**Acknowledgement**

The authors are deeply grateful to the high-level experts who contributed to the Delphi-inspired consultation.

**Declaration of generative AI and AI-assisted technologies**

During the preparation of this work the authors used Gemini 2.5 Pro for language editing of parts of the paper. After using this tool, the authors reviewed and edited the content as needed and takes full responsibility for the content of the publication.

# Appendix. Exploring the Potential of Citiverses for Regulatory Learning
## *Supplementary material*

*Supplementary Table 1. Comparison of experimentation spaces for regulatory learning (Adapted from Kert, Vebrova & Schade, 2022).*

| Feature | Test beds | Living labs | Regulatory sandboxes | Citiverses |
|---|---|---|---|---|
| **Goal** | Test technical performance and compliance of technologies in controlled environments. | Co-create and prototype solutions with user involvement in real-world settings. | Experiment with regulatory frameworks and innovations under controlled real-world conditions. | Provide immersive, large-scale virtual environments for experimentation with policies, technologies and societal impacts. |
| **Scalability** | Limited to specific technologies or infrastructure. | Constrained by geographical and community boundaries. | Focused on sector-specific or localized regulatory scenarios. | Highly scalable, supporting large-scale simulations of urban environments and policies across multiple domains and geographies. |
| **Cost-effectiveness** | High costs for infrastructure and equipment. | Moderate costs but limited to physical participation. | Moderate costs but can involve significant regulatory and operational complexities. | Cost-effective virtual simulations eliminate the need for physical infrastructure and real-world disruptions. |
| **Connected infrastructure** | Limited; primarily focused on isolated systems and components. | Relying on fragmented networks and lacking national-level connectivity. | Can test regulatory frameworks within specific contexts but lacks system-wide integration. | National cloud-based platforms with digital twin capabilities enable seamless connectivity and service delivery across cities of all sizes. |
| **Interoperability, cross-sectorial and cross-border testing** | Focused on specific technical standards; limited cross-domain testing. | Context-specific and not designed for interoperability testing across jurisdictions. | Limited cross-jurisdictional applications; focus on specific regulatory areas. | Support testing of interoperability standards and cross-border regulatory frameworks in a seamless digital environment. |
| **Feedback and iteration** | Data collected for analysis but lacks real-time feedback. | Integrate stakeholder feedback but not in real time. | Feedback loops depending on real-world timelines, slowing adaptation. | Real-time feedback and analytics enable faster iterations and learning. |

| Complex systems modelling | Simplistic technical testing with limited social or economic complexity. | Engage users but lack the ability to model complex urban interdependencies. | Focus on regulatory frameworks rather than systemic interactions. | Can integrate diverse datasets (e.g., environmental, social, economic) for holistic modelling of complex urban systems. |
|---|---|---|---|---|
| Citizen participation | Minimal, often limited to end-user testing. | Strong focus on user engagement but localized and small-scale. | Citizen involvement limited to affected groups; lack of proactive simulation capabilities. | Enable dynamic citizen participation through avatars, providing insights into social acceptance and unforeseen societal impacts in diverse scenarios. Enable visualisation of urban solutions and encourages inclusive participation in real-world challenges. |
| Risk mitigation, ethical and equity testing | Limited to technical risks; real-world testing carries inherent risks. | Focus on user co-creation but cannot test high-risk scenarios without real-world impact. | High-stakes scenarios can be tested but with real-world regulatory constraints. | Offer a safe, controlled virtual space to experiment with high-risk scenarios (e.g., public health crises, environmental disasters) and to address equity and ethical challenges without real-world consequences. |
| Application to emerging technologies | Focused on validating technical requirements. | Limited to early-stage prototyping; not suited for testing complex emerging technologies. | Allow some flexibility for testing regulatory implications of emerging technologies. | Ideal for integrated testing of technical, legal and ethical aspects of emerging technologies like AI and blockchain. AI- and IoT-powered tools enable urban planners to create tailored digital replicas of cities for precise planning and analysis. |

**Supplementary Figure 1.** *Expertise map: participants' self-assessed expertise levels in regulatory learning and virtual worlds, with citiverse expertise represented by circle size.*

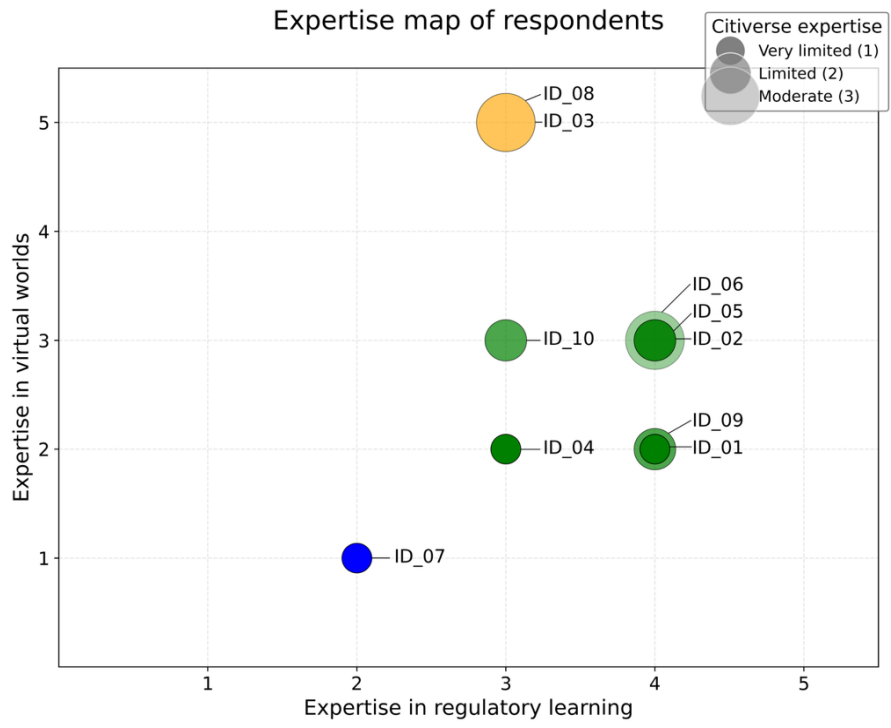